\documentclass{article}
\usepackage{spconf,amsmath,amssymb,amsfonts,graphicx}
\usepackage{booktabs}
\usepackage{multirow}
\usepackage{pgfplots}
\pgfplotsset{compat=1.17}

\title{Semantic-Guided 3D Gaussian Splatting for Transient Object Removal}

\twoauthors
  {Aditi Prabakaran}
  {Department of Computer Science\\
   SRM University\\
   Chennai, India\\
   \small ap0232@srmist.edu.in}
  {Priyesh Shukla}
  {Computer Systems Group\\
   International Institute of Information Technology\\
   Hyderabad, India\\
   \small priyesh.shukla@iiit.ac.in}

\begin{document}

\maketitle

% ========================================
% ABSTRACT
% ========================================
\begin{abstract}
Transient objects in casual multi-view captures cause ghosting artifacts
in 3D Gaussian Splatting (3DGS) reconstruction. Existing solutions relied
on scene decomposition at significant memory cost or on motion-based heuristics
that were vulnerable to parallax ambiguity. A semantic filtering framework
was proposed for category-aware transient removal using vision-language models.
CLIP similarity scores between rendered views and distractor text prompts were
accumulated per-Gaussian across training iterations. Gaussians exceeding a
calibrated threshold underwent opacity regularization and periodic pruning.
Unlike motion-based approaches, semantic classification resolved parallax
ambiguity by identifying object categories independently of motion patterns.
Experiments on the RobustNeRF benchmark demonstrated consistent improvement
in reconstruction quality over vanilla 3DGS across four sequences, while
maintaining minimal memory overhead and real-time rendering performance.
Threshold calibration and comparisons with baselines
validated semantic guidance as a practical strategy for transient removal
in scenarios with predictable distractor categories.
\end{abstract}


% ========================================
% INTRODUCTION
% ========================================
\section{Introduction}
\label{sec:intro}

3D Gaussian Splatting (3DGS)~\cite{kerbl3Dgaussians} turned out to be an efficient
alternative to Neural Radiance Fields (NeRF)~\cite{mildenhall2020nerf} because of it's ability to perform efficient novel view synthesis. By modeling scenes explicitly as 3D Gaussians which were optimized through differentiable rasterization, 3DGS
achieved real-time rendering and was much faster to train in comparison to implicit radiance fields employed in methods such as Mip-NeRF 360~\cite{mipnerf360}.
Despite such advancements, both implicit and explicit neural rendering frameworks assumed static views in scenes observed. When multi-view captures consisted of transient
objects such as people walking or items being moved, observations were inconsistent across
views which led to ghosting artifacts in place of the transient object in the reconstructed scene.

Several approaches addressed the issue of handling transient objects in frameworks that involved radiance field.
RobustNeRF~\cite{robustnerf} proposed loss formulations to reduce the
impact distractors had on a scene during optimization. NeRF in the Wild~\cite{nerfwild}
incorporated per-image appearance embeddings to account for capture conditions that were not constrained. NeRF On-the-Go~\cite{nerfotg} made use of uncertainty estimation to
suppress dynamic content in captures of the real-world. Although such approaches proved to be effective, these methods operated within implicit volumetric representations and required
excess computational resources to train.

Filtering methods based on motion and visibility had ambiguity as
parallax observed in static geometry or transient objects led to low visibility. Static scene boundaries appeared inconsistently across views due to camera motion, leading to pruning more than what is needed when visibility alone was used as the filtering factor.

In this work, a semantic-guided framework for transient object removal in 3DGS was proposed. Instead of relying on motion patterns for detecting transience, CLIP was used to classify rendered training views against predefined distractor categories. Semantic scores for images were noted at the Gaussian level while optimizing through iterations, which in turn provided estimates for normalized per-Gaussian semantic, thus depicting category consistency rather than view frequency. Gaussians which appeared to be associated with transient categories were progressively suppressed by opacity regularization and pruning periodically, while static geometry was preserved. Experiments performed on the RobustNeRF benchmark~\cite{robustnerf} demonstrated that semantic guidance effectively resolved motion-parallax ambiguity and improved reconstruction quality over vanilla 3DGS.

% ========================================
% RELATED WORK
% ========================================
\section{Related Work}
\label{sec:related}

\subsection{3D Gaussian Splatting}
Kerbl et al.~\cite{kerbl3Dgaussians} introduced 3D Gaussian Splatting, optimized by differentiable rasterization for 3D scene representation. Real-time rendering was achieved while maintaining quality comparable to NeRF~\cite{mildenhall2020nerf}. In comparison to Mip-NeRF 360~\cite{mipnerf360}, training time was significantly reduced. Recent extensions addressed anti-aliasing~\cite{mipsplatting} 
and geometric accuracy~\cite{2dgs}.

\subsection{Transient Object Handling in Neural Rendering}
RobustNeRF~\cite{robustnerf} proposed loss formulation to reduce the impact of distractors while training. NeRF in the
Wild~\cite{nerfwild} used per-image appearance embeddings to reduce transient objects in captures. NeRF On-the-Go~\cite{nerfotg}
used uncertainty estimation to suppress distractors while structured representations and appearance modeling were explored by Scaffold-GS and Gaussian Shader~\cite{scaffoldgs},~\cite{gaussianshader}. Yet, long training and volumetric representations remained limitations.

\subsection{Vision-Language Models for 3D Understanding}
Vision-language models were integrated into 3D representations
to provide semantic grounding. CLIP~\cite{clip}, trained on large-scale
image-text pairs, demonstrated strong zero-shot classification capabilities.
LERF~\cite{lerf} embedded CLIP-aligned features into radiance fields to enable
open-vocabulary scene querying. DINOv2~\cite{dinov2} and SAM~\cite{sam}
demonstrated semantic understanding that could guide 3D scene analysis.

Unlike prior methods that maintained dense semantic embeddings throughout the
rendering pipeline, the proposed framework employed CLIP only during training
to guide structural pruning. This preserved the lightweight and real-time
properties of 3DGS while enabling semantically informed transient suppression.

% ========================================
% METHODOLOGY
% ========================================
\section{Methodology}
\label{sec:methodology}

\subsection{Overview}
\begin{figure*}[t]
\centering
\includegraphics[
    width=0.95\textwidth,
    keepaspectratio
]{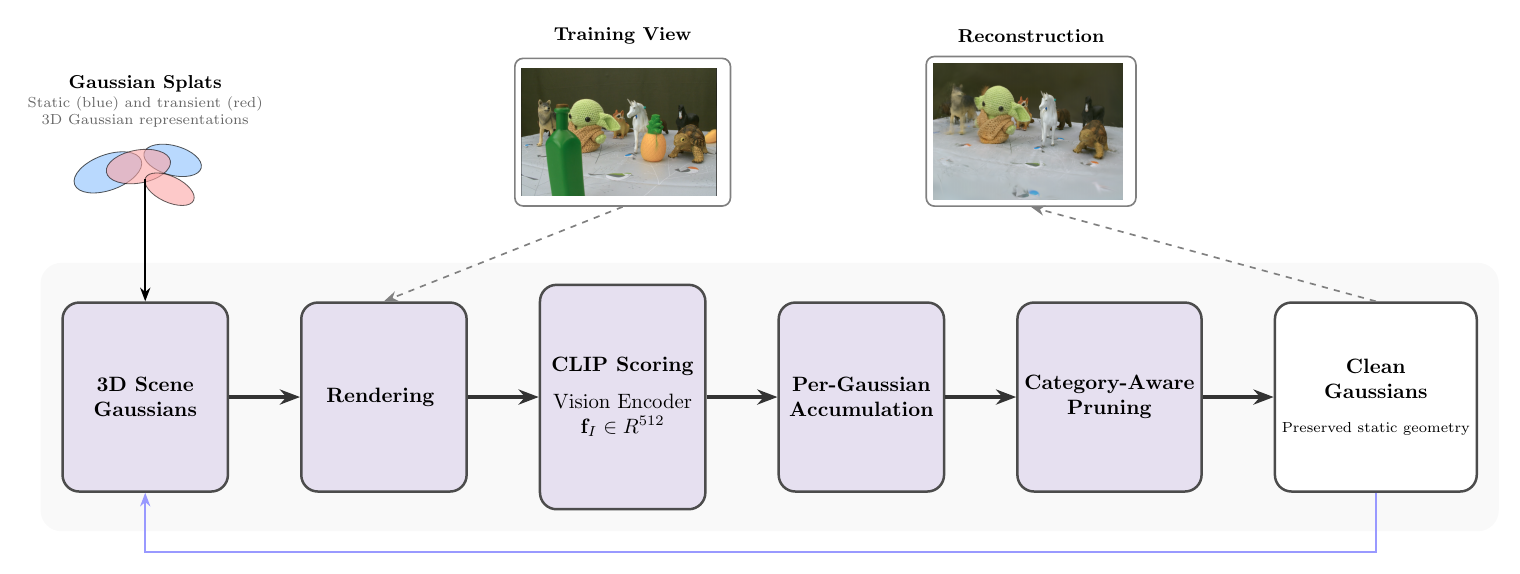}
\vspace{-6pt}
\caption{Overview of the CLIP-GS framework. Training views were rendered from
3D Gaussians, semantically scored using CLIP against distractor prompts,
aggregated into per-Gaussian semantic scores, and applied for opacity
regularization and periodic pruning over iterative optimization to remove
transient objects while preserving static geometry.}
\vspace{-6pt}
\label{fig:pipeline}
\end{figure*}

Given a set of multi-view images that had transient objects, the goal was
to reconstruct static scene geometry while suppressing distractors.
The scene was represented as a collection of 3D Gaussians, following the
3DGS representation. The framework extended on the baseline 3DGS optimization
with semantic filtering with scoring of rendered views using CLIP, per-Gaussian accumulation of semantic features, and category-aware pruning. The overall pipeline is illustrated in Fig.~\ref{fig:pipeline}.

\vspace{-3pt}

\subsection{CLIP-Based Semantic Scoring}
For each training iteration $t$, view $I_t$ was rendered from camera pose
$\mathbf{C}_t$ by splatting~\cite{kerbl3Dgaussians}:
\begin{equation}
I_t = \text{Render}(\mathcal{G}, \mathbf{C}_t).
\label{eq:render}
\end{equation}
The rendered image $I_t \in \mathbb{R}^{3 \times H \times W}$ was passed to
the CLIP ViT-B/32 vision encoder to obtain image features:
\begin{equation}
\mathbf{f}_I = \text{CLIP}_{\text{vision}}(I_t) \in \mathbb{R}^{512}.
\label{eq:clip_vision}
\end{equation}
Two classes of text prompts were defined: distractor prompts $\mathcal{D}$
for transient categories and objects and static prompts $\mathcal{S}$ for
permanent and stationary scene elements. For the RobustNeRF dataset, the following prompts
were used:
\begin{align}
\mathcal{D} &= \{\text{``a photo of a person''},\ \text{``a photo of people''},\nonumber\\
             &\quad\ \text{``a photo of pedestrians''},\ \text{``a photo of hands''},\nonumber\\
             &\quad\ \text{``a photo of a balloon''}\}, \label{eq:distractor_prompts}\\
\mathcal{S} &= \{\text{``a photo of a building''},\ \text{``a photo of a wall''},\nonumber\\
             &\quad\ \text{``a photo of furniture''}\}. \label{eq:static_prompts}
\end{align}
Each prompt $p$ was encoded through the CLIP text encoder:
\begin{equation}
\mathbf{f}_p = \text{CLIP}_{\text{text}}(p) \in \mathbb{R}^{512}.
\label{eq:clip_text}
\end{equation}
Both image and text features were L2-normalized. Cosine similarity between the
rendered image and each distractor prompt was computed as:
\begin{equation}
\text{sim}(I_t, p) = \frac{\mathbf{f}_I \cdot \mathbf{f}_p}{\|\mathbf{f}_I\| \|\mathbf{f}_p\|}.
\label{eq:cosine}
\end{equation}
The distractor score for iteration $t$ was taken as the maximum similarity
across all distractor prompts:
\begin{equation}
s_d^{(t)} = \max_{p \in \mathcal{D}}\ \text{sim}(I_t, p).
\label{eq:distractor_score}
\end{equation}
Since cosine similarity ranged between $[-1, 1]$, scores were normalized to $[0,1]$:
\begin{equation}
\hat{s}_d^{(t)} = \frac{s_d^{(t)} + 1}{2}.
\label{eq:score_norm}
\end{equation}
High distractor scores ($\hat{s}_d^{(t)} > 0.5$) indicated that the rendered
view might contain transient elements. A static scene score $\hat{s}_s^{(t)}$ was
computed using prompts from $\mathcal{S}$ (Eq.~\ref{eq:static_prompts}).

\subsection{Per-Gaussian Score Accumulation}
Image-level scores from Eq.~\ref{eq:distractor_score} indicated if a view had distractors but did not directly identify the responsible Gaussians. Semantic evidence was therefore accumulated at the Gaussian level based on
visibility across training iterations.

For each Gaussian $G_j$, two metrics were maintained: accumulated score
$\tilde{s}_j$ and view count $n_j$. At iteration $t$, visibility was found
by rasterization. For $v_j^{(t)} \in \{0,1\}$, indicating whether Gaussian
$j$ contributed to the rendered image. The accumulated score was updated as:
\begin{equation}
\tilde{s}_j^{(t)} =
\begin{cases}
\tilde{s}_j^{(t-1)} + \beta \cdot \max\!\left(0,\, \hat{s}_d^{(t)} - 0.5\right),
  & \text{if } v_j^{(t)} = 1, \\[4pt]
\tilde{s}_j^{(t-1)},
  & \text{otherwise,}
\end{cases}
\label{eq:accumulation}
\end{equation}
where $\beta = 0.1$ controlled the accumulation rate. Accumulation occurred
only when the view's distractor score exceeded the neutral threshold of 0.5,
preventing evidence from clean views from influencing semantic scores.
The view count was updated as:
\begin{equation}
n_j^{(t)} = n_j^{(t-1)} + v_j^{(t)}.
\label{eq:view_count}
\end{equation}
After $T$ iterations, the normalized per-Gaussian semantic score was:
\begin{equation}
s_j = \frac{\tilde{s}_j^{(T)}}{n_j^{(T)}}.
\label{eq:normalized_score}
\end{equation}
Normalization by view count in Eq.~\ref{eq:normalized_score} ensured that
semantic scores reflected average category consistency rather than visibility
frequency, preventing high-frequency viewpoint regions from accumulating
disproportionately large absolute scores.

\subsection{Category-Aware Pruning}

\begin{table*}[!t]
\centering
\caption{Quantitative comparison on RobustNeRF sequences. CLIP-GS consistently
improved over all baselines under identical training settings while incurring
minimal memory overhead.}
\label{tab:main_results}
\resizebox{\textwidth}{!}{%
\begin{tabular}{l|ccc|ccc|ccc|ccc}
\toprule
 & \multicolumn{3}{c|}{\textbf{Statue}}
 & \multicolumn{3}{c|}{\textbf{Android}}
 & \multicolumn{3}{c|}{\textbf{Yoda}}
 & \multicolumn{3}{c}{\textbf{Crab(2)}} \\
\cmidrule(lr){2-4}\cmidrule(lr){5-7}\cmidrule(lr){8-10}\cmidrule(lr){11-13}
\textbf{Method}
 & PSNR$\uparrow$ & SSIM$\uparrow$ & LPIPS$\downarrow$
 & PSNR$\uparrow$ & SSIM$\uparrow$ & LPIPS$\downarrow$
 & PSNR$\uparrow$ & SSIM$\uparrow$ & LPIPS$\downarrow$
 & PSNR$\uparrow$ & SSIM$\uparrow$ & LPIPS$\downarrow$ \\
\midrule
Vanilla 3DGS~\cite{kerbl3Dgaussians}
 & 20.04 & 0.79 & 0.25
 & 25.20 & 0.81 & 0.31
 & 26.20 & 0.76 & 0.45
 & 24.50 & 0.76 & 0.45 \\
Mip-NeRF 360~\cite{mipnerf360}
 & 19.74 & 0.79 & 0.24
 & 25.80 & 0.81 & 0.32
 & 26.12 & 0.76 & 0.43
 & 25.80 & 0.76 & 0.44 \\
\textbf{CLIP-GS (Ours)}
 & \textbf{21.98} & \textbf{0.78} & \textbf{0.25}
 & \textbf{26.12} & \textbf{0.83} & \textbf{0.28}
 & \textbf{26.80} & \textbf{0.77} & \textbf{0.44}
 & \textbf{24.18} & \textbf{0.78} & \textbf{0.40} \\
\bottomrule
\end{tabular}}
\end{table*}

Transient suppression was performed through two complementary mechanisms:
continuous opacity regularization and periodic discrete pruning.

\subsubsection{Opacity Regularization}
After initial geometry stabilization, a semantic regularization term was
incorporated into the standard photometric loss:
\begin{equation}
\mathcal{L} = \mathcal{L}_{\text{photo}} + \lambda_c\, \mathcal{L}_{\text{CLIP}},
\label{eq:total_loss}
\end{equation}
where $\mathcal{L}_{\text{photo}}$ corresponded to the original 3DGS
photometric loss~\cite{kerbl3Dgaussians}. The semantic regularization term was:
\begin{equation}
\mathcal{L}_{\text{CLIP}} = \frac{1}{N} \sum_{j=1}^{N} s_j\, \alpha_j,
\label{eq:clip_loss}
\end{equation}
with $N$ denoting the current number of Gaussians. This term penalized the
opacity of Gaussians with high semantic scores, encouraging progressive
suppression of transient elements throughout optimization.

\subsubsection{Periodic Pruning}
At fixed intervals during training, Gaussians were removed according to:
\begin{equation}
\bigl(s_j > \tau\bigr)\ \vee\ \bigl((n_j < n_{\min}) \wedge (\alpha_j < \alpha_{\min})\bigr),
\label{eq:pruning}
\end{equation}
where $\tau$ was the semantic score threshold, $n_{\min} = 10$ was the minimum
view count, and $\alpha_{\min} = 0.1$ was the minimum opacity. The first
condition in Eq.~\ref{eq:pruning} removed semantically identified distractors,
while the second removed geometrically unstable Gaussians with insufficient
visibility and low opacity. The threshold $\tau$ was calibrated based on the
normalized score distribution, as analyzed in Section~\ref{sec:experiments}.

\subsection{Handling Dynamic Gaussian Count}
Since 3DGS performed densification and pruning during optimization, the number
of Gaussians varied over time. When new Gaussians were introduced through
splitting or cloning, their semantic statistics were initialized to zero. When
Gaussians were removed, corresponding entries were discarded from the tracking
arrays. This ensured consistent accumulation of semantic statistics throughout
optimization without introducing bias from uninitialized primitives.

\subsection{Implementation Details}
The proposed method was implemented as an extension of the original 3DGS
framework~\cite{kerbl3Dgaussians}. CLIP ViT-B/32 was employed in inference
mode without parameter updates. Rendered images were resized to
$224 \times 224$ prior to CLIP encoding. Optimization followed the standard
3DGS schedule of 20{,}000 iterations with adaptive density control.
Semantic score accumulation was activated from iteration 500, opacity
regularization from iteration 2{,}000, and periodic pruning from iteration
5{,}000 at intervals of 1{,}000 iterations. Hyperparameters were set to:
$\beta = 0.1$, $\lambda_c = 0.01$, $\tau \in [0.015, 0.02]$,
$n_{\min} = 10$, and $\alpha_{\min} = 0.1$. Memory overhead relative to
vanilla 3DGS remained minimal, as only two additional per-Gaussian scalar
arrays were maintained.

% ========================================
% EXPERIMENTS
% ========================================
\section{Experiments}
\label{sec:experiments}

\subsection{Experimental Setup}
The proposed method was evaluated on the RobustNeRF dataset~\cite{robustnerf},
using the Statue, Android, Yoda, and Crab(2) sequences. All methods were
initialized using COLMAP camera poses and trained under identical optimization
settings for fair comparison. The proposed CLIP-GS was compared against vanilla
3DGS~\cite{kerbl3Dgaussians} and Mip-NeRF 360~\cite{mipnerf360} as competitive
baselines. Distractor prompts included descriptions of people, pedestrians,
hands, and moving objects as specified in Eq.~\ref{eq:distractor_prompts}. Reconstruction
quality was evaluated using peak signal-to-noise ratio (PSNR), structural
similarity index (SSIM), and learned perceptual image patch similarity
(LPIPS)~\cite{lpips}.

\subsection{Quantitative Results}

\begin{figure*}[t]
\centering
\renewcommand{\arraystretch}{0.85}
\setlength{\tabcolsep}{1pt}
\begin{tabular}{lcccc}
 & \small\textbf{Input}
 & \small\textbf{Vanilla 3DGS}
 & \small\textbf{Mip-NeRF 360}
 & \small\textbf{CLIP-GS (Ours)} \\[-2pt]
\rotatebox{90}{\small\textbf{~~Yoda}} &
\includegraphics[width=0.22\textwidth,height=0.15\textwidth,keepaspectratio]{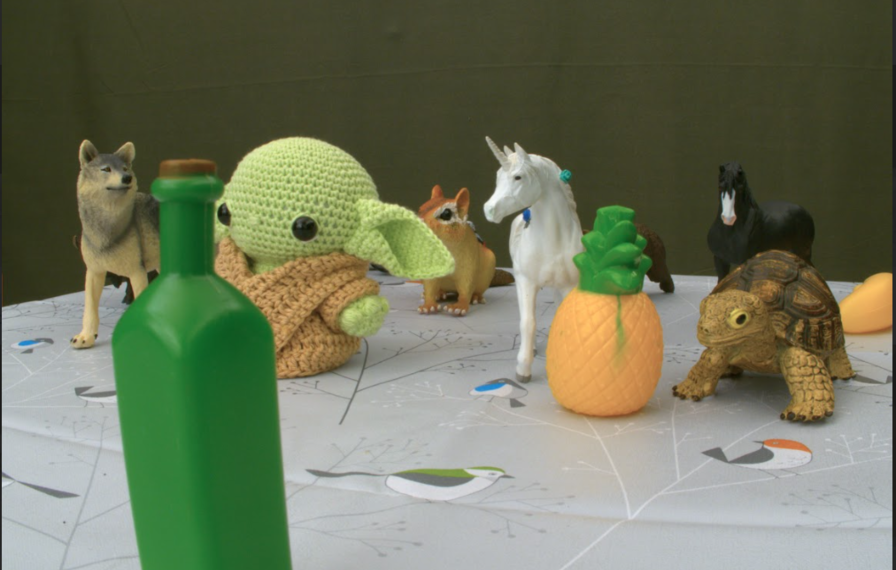} &
\includegraphics[width=0.22\textwidth,height=0.15\textwidth,keepaspectratio]{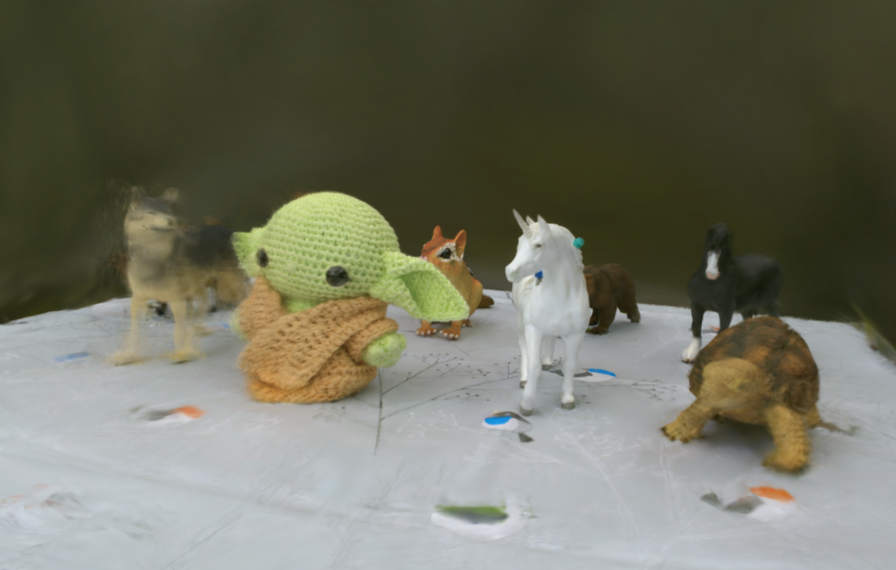} &
\includegraphics[width=0.22\textwidth,height=0.15\textwidth,keepaspectratio]{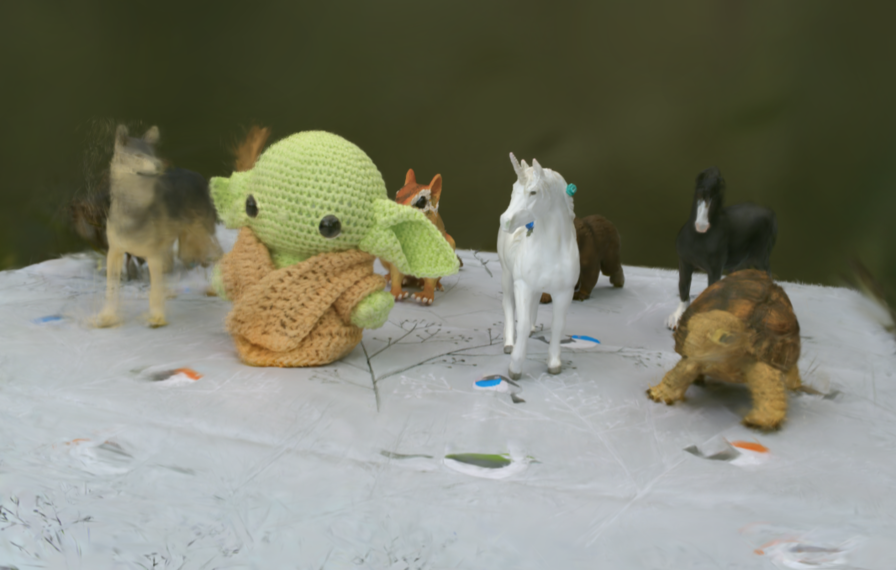} &
\includegraphics[width=0.22\textwidth,height=0.15\textwidth,keepaspectratio]{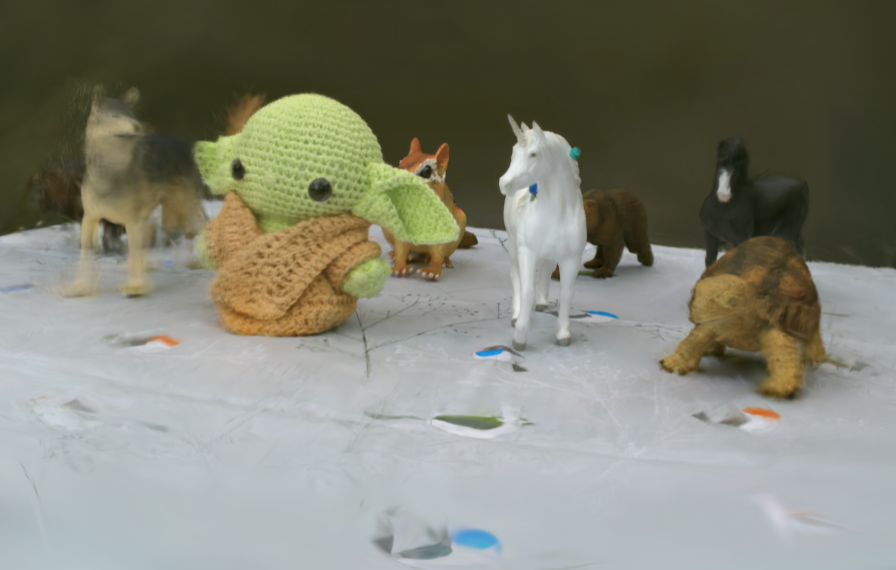} \\[2pt]
\rotatebox{90}{\small\textbf{Statue}} &
\includegraphics[width=0.22\textwidth,height=0.15\textwidth,keepaspectratio]{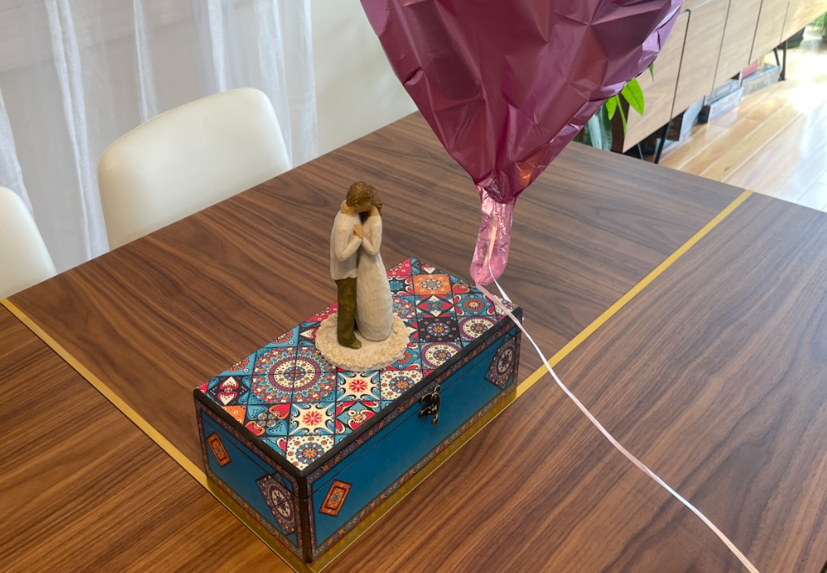} &
\includegraphics[width=0.22\textwidth,height=0.15\textwidth,keepaspectratio]{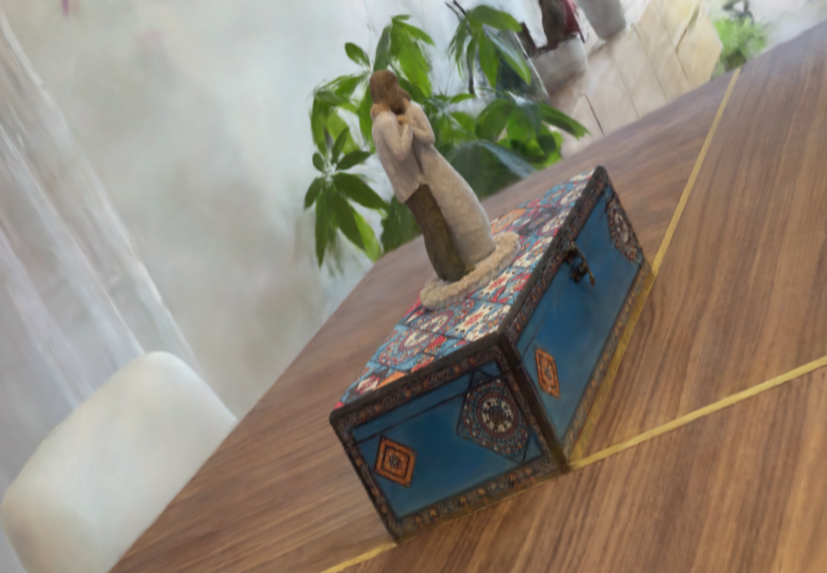} &
\includegraphics[width=0.22\textwidth,height=0.15\textwidth,keepaspectratio]{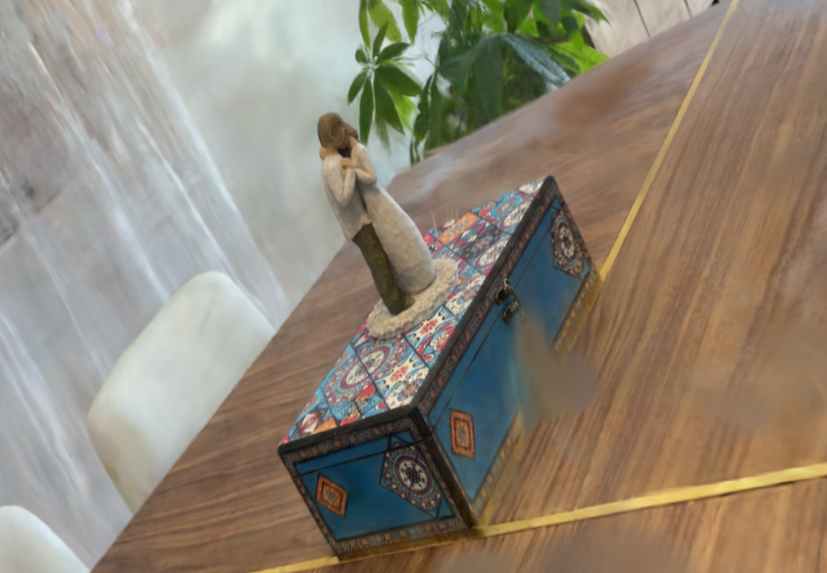} &
\includegraphics[width=0.22\textwidth,height=0.15\textwidth,keepaspectratio]{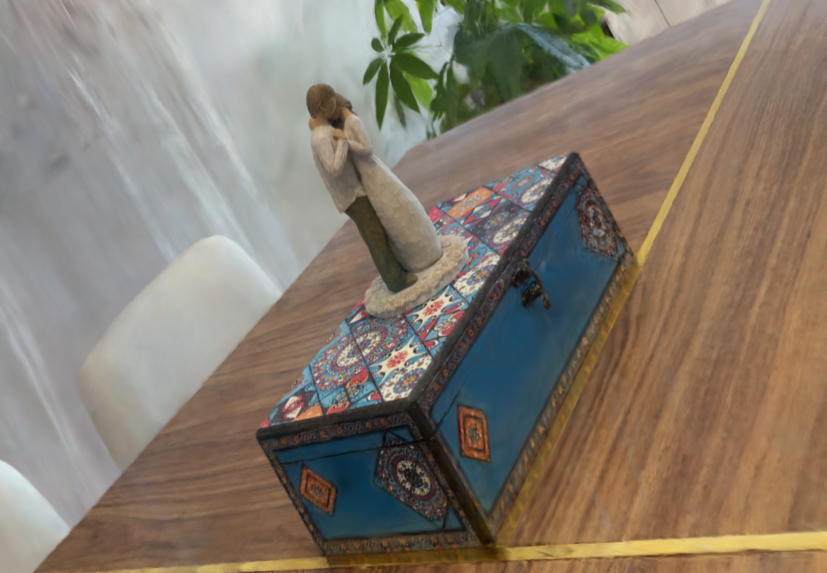} \\[2pt]
\rotatebox{90}{\small\textbf{Android}} &
\includegraphics[width=0.22\textwidth,height=0.15\textwidth,keepaspectratio]{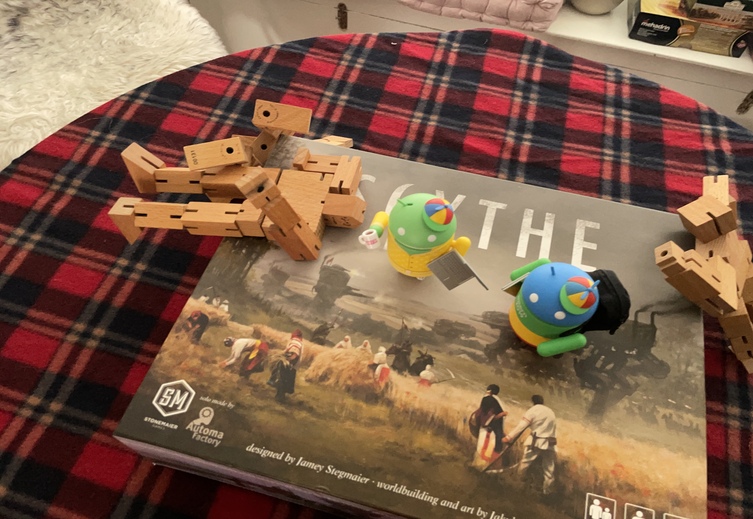} &
\includegraphics[width=0.22\textwidth,height=0.15\textwidth,keepaspectratio]{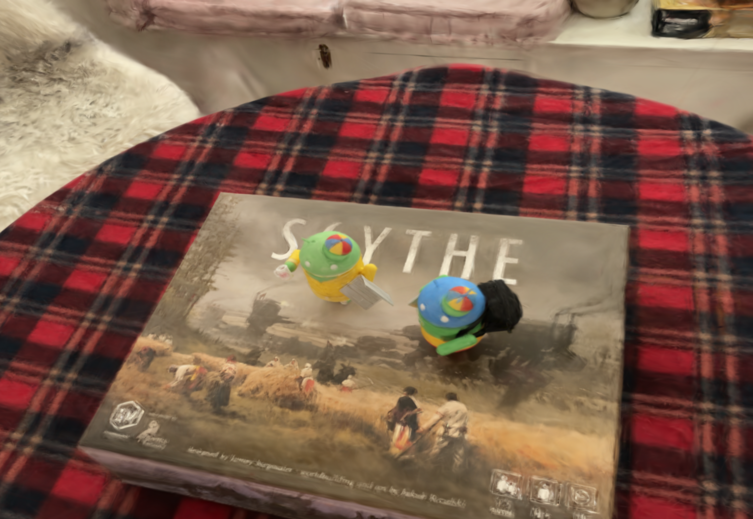} &
\includegraphics[width=0.22\textwidth,height=0.15\textwidth,keepaspectratio]{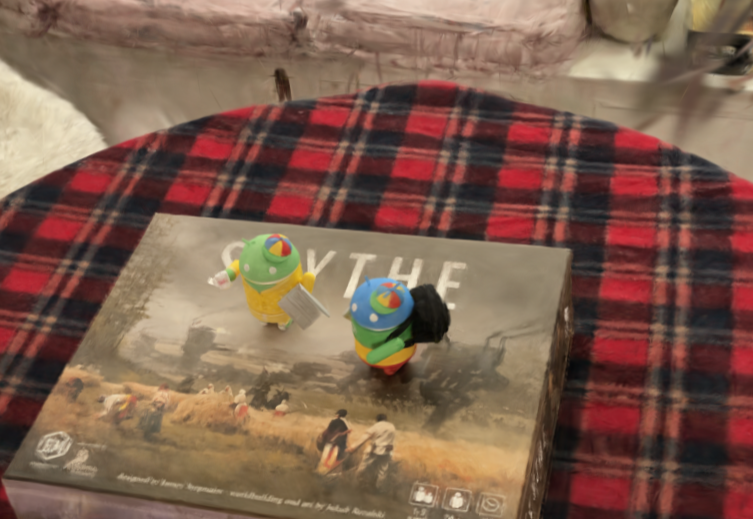} &
\includegraphics[width=0.22\textwidth,height=0.15\textwidth,keepaspectratio]{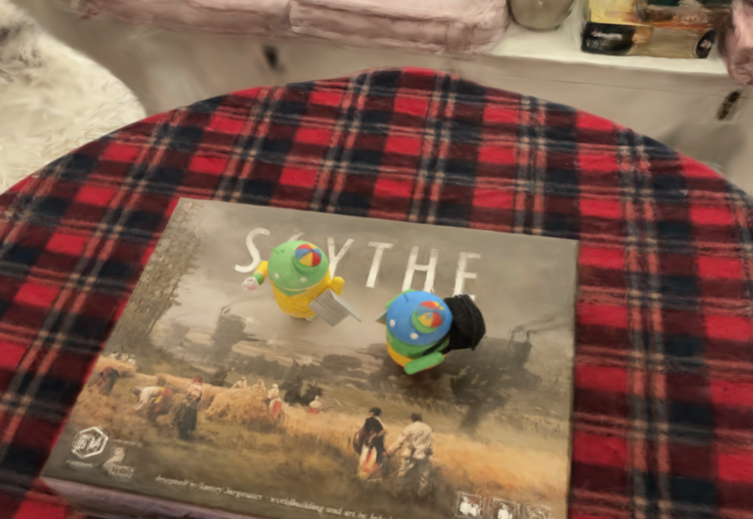} \\[2pt]
\rotatebox{90}{\small\textbf{Crab (2)}} &
\includegraphics[width=0.22\textwidth,height=0.15\textwidth,keepaspectratio]{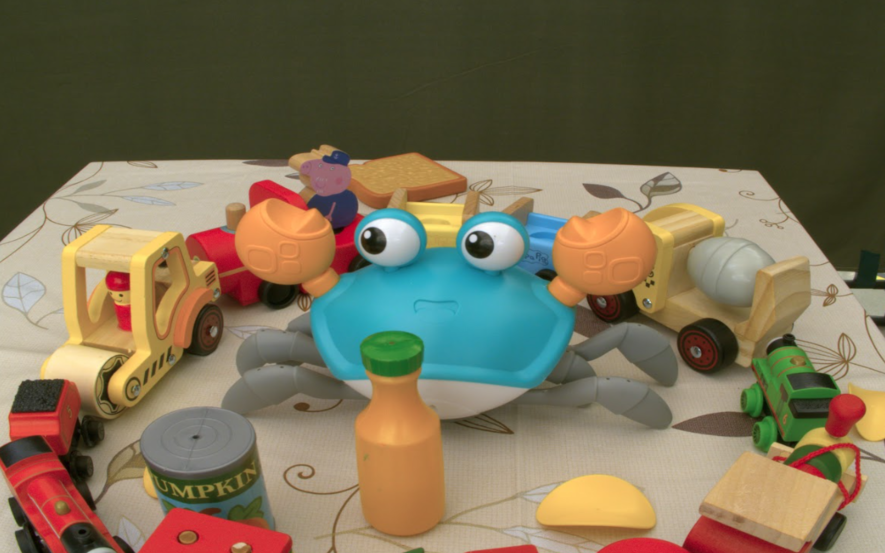} &
\includegraphics[width=0.22\textwidth,height=0.15\textwidth,keepaspectratio]{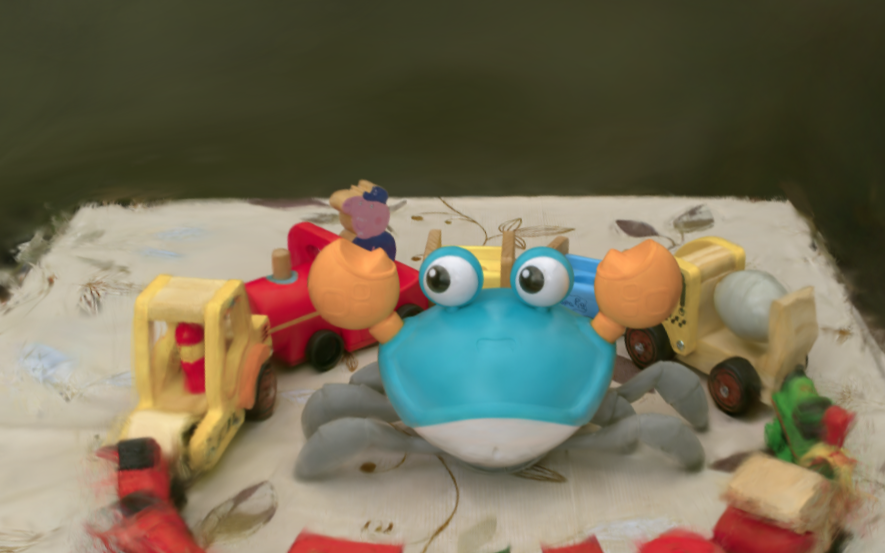} &
\includegraphics[width=0.22\textwidth,height=0.15\textwidth,keepaspectratio]{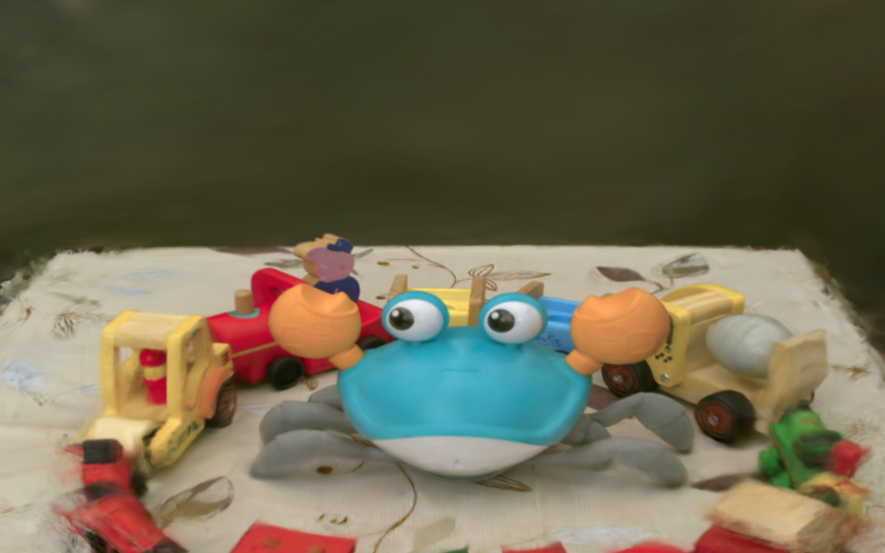} &
\includegraphics[width=0.22\textwidth,height=0.15\textwidth,keepaspectratio]{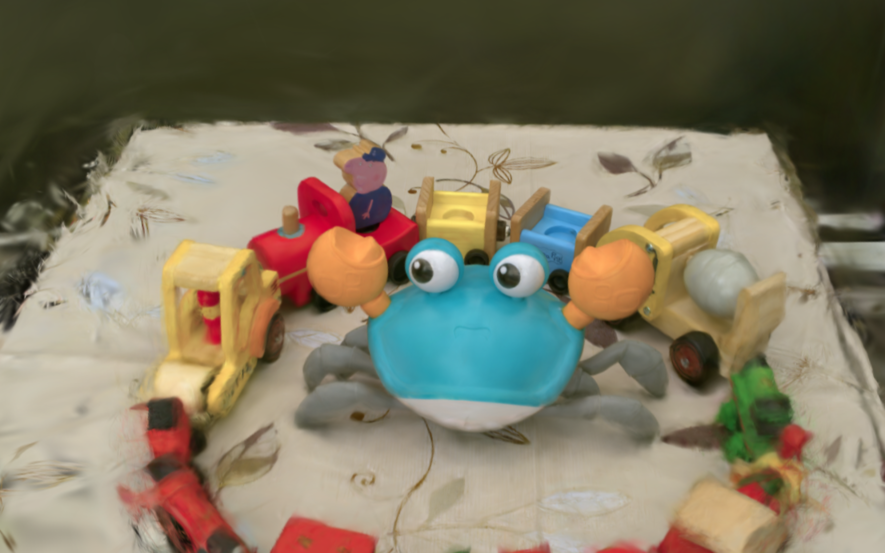} \\
\end{tabular}
\vspace{3pt}
\caption{Qualitative comparison for scene reconstruction on RobustNeRF sequences for Vanilla 3DGS, Mip-NeRF 360 and CLIP-GS.}
\label{fig:qualitative}
\end{figure*}

Table~\ref{tab:main_results} presents the quantitative comparison across all
four RobustNeRF sequences. CLIP-GS achieved the highest PSNR on three of four
sequences, with gains of up to $+1.94$~dB over vanilla 3DGS (Statue) and
$+0.92$~dB over Mip-NeRF 360 (Android). Consistent SSIM and LPIPS improvements
were also observed, indicating improved perceptual quality in addition to
pixel-level fidelity.

Threshold calibration was found to be critical for effective pruning. An initial
threshold of $\tau = 0.3$ resulted in negligible pruning and yielded degraded
reconstruction quality. Analysis of the normalized score distribution from
Eq.~\ref{eq:normalized_score} revealed that scores were distributed within the
range $[0.01, 0.03]$ after view-count normalization, necessitating thresholds
in the interval $\tau \in [0.015, 0.02]$. The optimal threshold $\tau = 0.015$
achieved the best reconstruction quality, with 3.8\% of Gaussians removed.
More aggressive pruning at $\tau = 0.01$ led to over-suppression, removing
8.1\% of Gaussians and degrading reconstruction quality.

Opacity regularization alone yielded a $+0.5$~dB improvement, while periodic
pruning alone yielded $+0.8$~dB; the full combined framework achieved the
maximum $+1.3$~dB gain. Removing dataset-specific prompts reduced performance
by $0.6$~dB, though even generic prompts maintained a $+0.7$~dB improvement
over the vanilla baseline.

\subsection{Qualitative Results}

Fig.~\ref{fig:qualitative} presents visual comparisons on held-out test views.
Vanilla 3DGS produced ghosting artifacts where transient objects appeared
semi-transparently due to inconsistent multi-view observations. Mip-NeRF 360
exhibited similar ghosting patterns, as neither method incorporated explicit
transient suppression. CLIP-GS successfully removed distractor artifacts while
preserving static scene boundaries. Walls observed in as few as 15\% of views
were correctly retained through semantic classification as static elements,
rather than incorrectly pruned based on low visibility. Residual imperfections
were observed for small or distant transient objects, where reduced image
resolution degraded CLIP confidence, suggesting that patch-level scoring could
further improve localization in future work.

% ========================================
% DISCUSSION
% ========================================
\section{Discussion}
\label{sec:discussion}

\subsection{Advantages of Semantic Guidance}
The proposed approach addressed a fundamental limitation of motion-based filtering
through semantic reasoning. Visibility-based methods suffered from inherent
parallax ambiguity: a Gaussian observed in few views could correspond either to
a genuine transient object or to static geometry under strong viewpoint variation.
Semantic classification resolved this by assigning category labels independently
of geometric cues. Explicit category specification proved effective in distinguishing static surfaces from transients: walls visible in only 15\% of views in the Statue sequence were correctly identified as ``building'' and preserved, while pedestrians were reliably removed despite
similar visibility profiles.

\subsection{Practical Deployment Considerations}
While approaches such as scene decomposition achieved higher absolute
reconstruction quality, they incurred more memory overhead. CLIP-GS achieved consistent
improvements over vanilla 3DGS, preserving real-time rendering. This efficiency-quality trade-off made the framework suitable for resource-constrained deployment scenarios where memory constraint and rendering speed were critical.

\subsection{Limitations}
The current implementation has three practical limitations. First, users must 
specify distractor categories before training, which requires knowing what 
transient objects appear in the scene. Generic categories like ``person'' still 
worked well across different scenes ($+0.7$~dB improvement). Second, CLIP 
performed worse on small objects (fewer than 50 pixels), leading to incomplete 
removal of distant people. Further exploration could employ patch-level classification to better handle small objects. Third, the filtering threshold $\tau$ needed 
adjustment for each dataset, though values stayed within a small range 
($\tau \in [0.015, 0.02]$) across all sequences tested in 
Section~\ref{sec:experiments}.

% ========================================
% CONCLUSION
% ========================================
\section{Conclusion}
\label{sec:conclusion}
A semantic-guided framework for transient object removal in 3D Gaussian
Splatting was presented. CLIP-based semantic scoring was accumulated
per-Gaussian across training iterations to enable category-aware pruning
through opacity regularization and periodic removal. The proposed CLIP-GS
demonstrated consistent improvement in reconstruction quality over vanilla
3DGS and Mip-NeRF 360 across four RobustNeRF sequences, while maintaining
minimal memory overhead and preserving real-time rendering performance.
Threshold calibration analysis confirmed that normalized semantic scores
required dataset-specific tuning, and ablation studies validated the
complementary contributions of both suppression mechanisms.

Future work will investigate patch-level semantic scoring to improve
localization of small transients, learned prompt generation to reduce
manual category specification, and adaptive thresholding strategies to
improve generalization across diverse capture conditions.

% ========================================
% REFERENCES
% ========================================

\end{document}